\documentclass[10pt,twocolumn,letterpaper]{article}

\usepackage{cvpr}              

\usepackage{graphicx}
\usepackage{amsmath}
\usepackage{amssymb}
\usepackage{booktabs}
\usepackage{bm}
\usepackage{pifont}

\usepackage[pagebackref=true,breaklinks=true,letterpaper=true,colorlinks,bookmarks=false]{hyperref}

\usepackage{marvosym}
\usepackage{balance}
\usepackage{arydshln}
\usepackage{multirow}
\usepackage{makecell}

\newcommand{\tablestyle}[2]{\setlength{\tabcolsep}{#1}\renewcommand{\arraystretch}{#2}\centering\footnotesize}
\newlength\savewidth\newcommand\shline{\noalign{\global\savewidth\arrayrulewidth
		\global\arrayrulewidth 1pt}\hline\noalign{\global\arrayrulewidth\savewidth}}
\usepackage[table]{xcolor}

\usepackage[capitalize]{cleveref}
\crefname{section}{Sec.}{Secs.}
\Crefname{section}{Section}{Sections}
\Crefname{table}{Table}{Tables}
\crefname{table}{Tab.}{Tabs.}

\definecolor{Gray}{gray}{0.5}

\definecolor{Highlight}{HTML}{39b54a}  
\definecolor{Modify}{HTML}{2240F0}  

\definecolor{Highlight}{HTML}{3DA600}  

\newcommand{\cmark}{\color{darkgray}\ding{51}}%
\newcommand{\xmark}{\color{darkgray}\ding{55}}%

\newcommand*{\affaddr}[1]{#1} 
\newcommand*{\affmark}[1][*]{\textsuperscript{#1}}

\makeatletter
\newcommand{\printfnsymbol}[1]{\textsuperscript{\@fnsymbol{#1}}}
\makeatother

\def\cvprPaperID{7298} 
\def\confName{CVPR}
\def\confYear{2022}

\makeatletter
\def\thanks#1{\protected@xdef\@thanks{\@thanks
        \protect\footnotetext{#1}}}
\makeatother
\begin{document}

\title{Bridged Transformer for Vision and Point Cloud 3D Object Detection}

\author{
	Yikai Wang\affmark[1]\quad$\;$ TengQi Ye\affmark[2]\quad$\;$ Lele Cao\affmark[1]\quad$\;$ Wenbing Huang\affmark[3]\quad$\;$\\\vspace{0.05in}Fuchun Sun\affmark[1]\textsuperscript{\Letter}$\thanks{\textsuperscript{\Letter}~Corresponding author: Fuchun Sun.}$\quad$\;$ Fengxiang He\affmark[4]\quad$\;$ Dacheng Tao\affmark[4]\\
	\affaddr{\affmark[1]Beijing National Research Center for Information Science and Technology$\,$(BNRist),\\ State Key Lab on Intelligent Technology and Systems,\\ Department of Computer Science and Technology, Tsinghua University}\quad \affaddr{\affmark[2]ByteDance Inc.}
	\\\affaddr{\affmark[3]Institute for AI Industry Research (AIR), Tsinghua University}\quad\affaddr{\affmark[4]JD Explore Academy, JD.com}\\
	\tt\small{wangyk17@mails.tsinghua.edu.cn, yetengqi@gmail.com, caolele@gmail.com, hwenbing@126.com, }\\
	\tt\small{fuchuns@tsinghua.edu.cn, fengxiang.f.he@gmail.com, dacheng.tao@gmail.com}\\
}

\maketitle

\begin{abstract}
3D object detection is a crucial research topic in computer vision, which usually uses 3D point clouds as input in conventional setups. Recently, there is a trend of leveraging multiple sources of input data, such as complementing the 3D point cloud with 2D images that often have richer color and fewer noises. However, due to the heterogeneous geometrics of the 2D and 3D representations, it prevents us from applying off-the-shelf neural networks to achieve multimodal fusion. To that end, we propose Bridged Transformer (BrT), an end-to-end architecture for 3D object detection. BrT is simple and effective, which  learns to identify 3D and 2D object bounding boxes from both points and image patches. A key element of BrT lies in the utilization of object queries for bridging 3D and 2D spaces, which unifies different sources of data representations in Transformer. We adopt a form of feature aggregation realized by point-to-patch projections which further strengthen the correlations between images and points. Moreover, BrT works seamlessly for fusing the point cloud with multi-view images. We experimentally show that BrT surpasses state-of-the-art methods on SUN RGB-D and ScanNetV2 datasets. 

\end{abstract}

\section{Introduction}
\label{sec:intro}

3D object detection, which aims at identifying or locating objects in 3D scenes, is drawing increasing attention and is acting as a fundamental task towards scene understanding. Many successful attempts~\cite{qi2019deep,xie2020mlcvnet,chen2020hierarchical,liu2021group} have been made using point cloud data as input. These attempts include converting the points to regular format (e.g., 3D voxel grids \cite{wu20153d}, polygon meshes \cite{kokkinos2012intrinsic}, multi-views \cite{su2015multi}), 
or using 3D specific operators (e.g., symmetric functions \cite{qi2017pointnet}, voting \cite{qi2019deep}) to design grouping strategies for points. In addition, since Transformers could be naturally permutation invariant and capable of capturing large-scale data correlations, they are lately applied to 3D object detection and  demonstrate superior performance \cite{misra2021end, liu2021group}. Besides handling point cloud learning tasks, Transformers have swept across various 2D tasks, e.g., image classification \cite{dosovitskiy2020image,liu2021swin}, object detection\cite{zhu2020deformable,fang2021you,carion2020end}, and semantic segmentation \cite{SETR, xie2021segformer}.

Deep multimodal learning by leveraging the advantage of  multiple modalities  has shown its superiority on various applications~\cite{journals/pami/BaltrusaitisAM19,DBLP:conf/nips/WangHSXRH20}. Despite the success of Transformers in 2D or 3D single-modal object detection tasks, 
the attempt of combining advantages from both point clouds and images   remains scarce. For 3D learning tasks, the point cloud provides essential geometrical cues, while the information in rich color images can  complement the point cloud by fulfilling the missing color information and correcting  noise errors. As a result,  the performance of 3D object detection could be potentially improved by the involvement of 2D images. One intuitive method is to lift 3-dimensional RGB vectors from images to extend the point features. A CNN-based 3D detection model, imVoteNet~\cite{qi2020imvotenet}, points out the difficulty in migrating 2D/3D discrepancies by this intuitive method, and instead, imVoteNet substitutes the RGB vectors with  image features extracted by a pre-trained 2D detector. However, simultaneously relying on both the image voting and point cloud voting assumptions in ~\cite{qi2020imvotenet} could accumulate the intrinsic grouping errors as mentioned by~\cite{liu2021group}. To avoid the learning process of point clouds  being impacted by middle-level 2D/3D feature interactions, \cite{qi2020imvotenet}  combines multimodal features over the first  layer, which potentially prevents the network from fully exploiting their semantic correlations or migrating multimodal discrepancies.

In this work, we propose Bridged Transformer (BrT) -- a simple and effective Transformer framework for 3D object detection. BrT bridges the learning processes of images and point clouds inside Transformer. This approach takes the sampled points and image patches as input. To protect the self-learning process of each modality, attentions between point tokens and image patch tokens are blocked but correlated by object queries throughout the Transformer layers. To strengthen the correlations of images and points, BrT is also equipped with powerful bridging designs from two perspectives. Firstly, we leverage conditional object queries for images and points that are  aware of the learned proposal points. Such design together with aligned  positional embeddings tells Transformer that object quires of images and points are aligned. Secondly, despite the perspective from object queries, we perform point-to-patch projections to explicitly leverage the spatial relationships of  both modalities. BrT  avoids the grouping errors due to its natural ability of capturing
long-range dependencies and global contextual information, and instead of lifting image features to point clouds at the beginning layer in~\cite{qi2020imvotenet}, BrT allows the full propagation of feature interactions in the whole network. As an additional advantage, BrT can be extended to combine point clouds with  multi-view images.

We evaluate BrT on both SUN RGB-D and ScanNetV2 datasets, where respectively, BrT achieves remarkably 2.4\% and 2.2\% improvements over state-of-the-art methods. 

To summarize, the contributions of our work are:
\begin{itemize}
    \item We propose BrT,  a novel framework for 3D object detection that bridges the learning processes of images and point clouds inside Transformer. 
    \item We propose to strengthen the correlation of images and points from two bridging perspectives including conditional object queries and the point-to-patch projection.
    \item BrT achieves the state-of-the-art on two benchmarks, which demonstrates the superiority of our design and also the potential in multi-view scenarios.
\end{itemize}

\section{Related Work}\label{sec:related-work}

\textbf{3D  detection with point cloud.} There are unique challenges faced by the processing of point clouds using deep neural networks (DNNs)~\cite{qi2019deep,xie2020mlcvnet,chen2020hierarchical,zhang2020h3dnet,liu2021group,misra2021end}. A detailed discussion around this difference can be found in~\cite{guo2020deep}. The targets of object detection in 3D space are locating 3D bounding boxes and recognizing the object classes. 
VoxelNet \cite{zhou2018voxelnet} proposes to divide a point cloud into equally spaced 3D voxels, and then transforms the points in each voxel into a unified feature representation. 
VoteNet \cite{qi2019deep} reformulates Hough voting in the context of deep learning to generate better points for box proposals with grouping. Transformers are also adapted to become suitable for handling 3D points. 3DTR~\cite{misra2021end} introduces an end-to-end Transformer with non-parametric queries and Fourier positional embeddings. 
Group-Free~\cite{liu2021group} adopts the attention mechanism to learn the point features, which potentially retains the information of all points to avoid the errors of previous grouping strategies.
Voxel Transformer~\cite{mao2021voxel} effectively captures the long-range relationships between voxels.

\vspace{0.05in}
\textbf{3D detection with multimodal data.} There are a few works that use deep networks to combine point clouds and images. MV3D~\cite{chen2017multi} proposes an element-wise fusion of representations from different domains,  based on the rigid assumption that all objects are on the same spatial plane and can be pinpointed solely from a top-down view of the point cloud. PointFusion~\cite{xu2018pointfusion} concatenates point cloud features and image features at two different levels to  learn their correlations, which could not guarantee the alignment of features. ImVoteNet~\cite{qi2020imvotenet} lifts crafted semantic and texture features to the 3D seed points for fusion. However, ImVoteNet is still negatively affected by the errors of grouping and combining features only at the beginning layer, leading to highly restricted feature interactions. Different from the aforementioned methods, our BrT fully exploits the feature correlation for images and points with additional bridging processes to strengthen the correlation.

\textbf{Transformer for 2D detection.} Recently, Transformer achieves the cutting edge performance in computer vision tasks~\cite{dosovitskiy2020image,carion2020end,liu2021swin,zhu2020deformable,fang2021you,yuan2021hrformer,meng2021conditional}.
For 2D object detection based on images, DETR~\cite{carion2020end} enables the Transformer to learn relations of the objects and the global image context to directly output the final set of predictions; and it also removes the need for non-maximum suppression and anchor generation. With the help of pre-training, YOLOS ~\cite{fang2021you} proposes a pure sequence-to-sequence approach that achieves competitive performance for object detection; hence it also tackles the transferability of Transformer from image recognition to object detection. 
Deformable DETR~\cite{zhu2020deformable} is an efficient and fast-converging model with attention modules only paying attention to a small set of tokens instead of the whole contexts. Conditional DETR~\cite{meng2021conditional} learns a conditional spatial query aiming to accelerate the training process.

\section{Method}
In this section, we propose Bridged Transformer (BrT) for 3D object detection with both vision and point cloud as input. 
We describe the overall structure of BrT in \cref{sec:architecture},
followed by the design of building blocks in \cref{sec:building-block}. We consider two aspects to bridge the learning processes of vision and point cloud in \cref{sec:bridge_by_query} and \cref{sec:alignment-projection}, respectively.

\begin{figure*}[t!]
\centering
\hskip-0.1in
\includegraphics[scale=0.57]{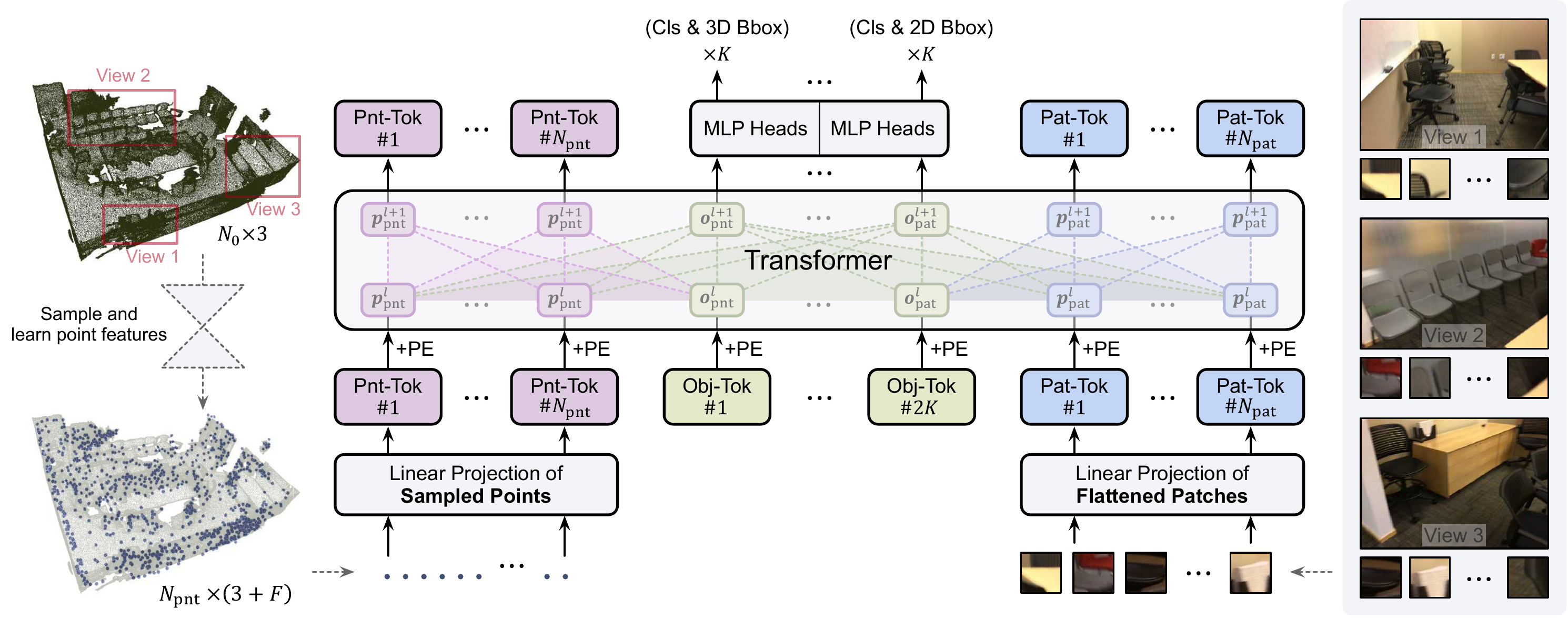}
\caption[]{Overall architecture of our Bridged Transformer (BrT) for  3D object detection based on point clouds and single-view/multi-view images. For each image view, we annotate its corresponding region on the point cloud for better readability. }
\label{pic:framework}
\end{figure*}

\subsection{Overall architecture}
\label{sec:architecture}
An overall architecture of our BrT is depicted in \cref{pic:framework}.
Suppose we are given $N\times 3$ points representing the 3D coordinates, and an $H\times W \times 3$ image. Here, $N$ is the number of points; $H$ and $W$ are the height and width of the image respectively.
For simplicity, we first analyze one image per scene since it matches the common scenario where the camera sensors capture (depth) points and RGB at the same time. 
Yet our method can be extended to handle multiple images per scene with different views, at one's disposal, as described in \cref{sec:multiview} and evaluated by our experiments.

Before feeding the {\bf point cloud data} to the first Transformer stage\footnote{Here, each stage contains a multi-head self-attention, an MLP, and two layer normalizations.},
we process the data with the method adopted in~\cite{qi2019deep}.
Specifically, we first sample $N_{\mathrm{pnt}}\times(3+F)$ ``seed points'' from a total of $N_0\times 3$ points, using PointNet++. 
Note that $N_{\mathrm{pnt}}$ denotes the number of sampled points; 
The positive integers $3$ and $F$ represent the dimension of the 3D Euclidean coordinate and point feature, respectively. 

For processing the {\bf image data}, we follow some successful practices from vision Transformers. 
Concretely, each image is evenly partitioned into $N_{\mathrm{pat}}$ patches before embedded by a multi-layer perception (MLP). 
Together with the embedded images patches, the learned object queries are sent to the model, generating output embeddings that are used to predict box coordinates and class labels.

Moreover, we adopt $2K$ learnable object queries, among which $K$ queries for points and $K$ for  image patches.
In summary, we have $N_{\mathrm{pnt}}+N_{\mathrm{pat}}$ basic tokens and $2K$ object queries tokens. 
Suppose the hidden dimension is $D$, The token features fed to the $l$-th ($l=1,\ldots,L$) Transformer stage contains 
point tokens $\bm{p}_{\mathrm{pnt}}^{l}\in\mathbb{R}^{N_{\mathrm{pnt}}\times D}$, 
patch tokens $\bm{p}_{\mathrm{pat}}^{l}\in\mathbb{R}^{N_{\mathrm{pat}}\times D}$, 
object queries for points $\bm{o}_{\mathrm{pnt}}^{l}\in\mathbb{R}^{K\times D}$, 
and object queries for patches $\bm{o}_{\mathrm{pat}}^{l}\in\mathbb{R}^{K\times D}$.

When given camera intrinsic and extrinsic parameters, each 3D point could be projected to the camera plane, 
that is correlating the 3D coordinates with 2D image pixels. 
We define the projection operator $\mathrm{proj}: \mathbb{R}^{3}\to\mathbb{R}^{2}$ indicating the projection process from a 3D point coordinate $\bm{k}=[x,y,z]^\top$ to a 2D pixel coordinate $\bm{k}'=[u,v]^\top$ on the corresponding image, and there is
\begin{align}
\label{eq:projection}
\bm{k}' =
\mathrm{proj}(\bm{k})=
\bm{\Pi}
\left[ \begin{array}{ccc}
\frac{1}{4} & 0 & 0 \\
0 & \frac{1}{4} & 0 \\
0 & 0 & 1
\end{array} \right] \bm{K} \bm{R}_t
\left[\begin{array}{c} x \\ y \\ z \\ 1 \end{array}\right],
\end{align}
where $\bm{K}$ and $\bm{R}_t$ are the intrinsic and extrinsic matrices, and $\bm{\Pi}$ is a perspective mapping.

BrT has $2K$ outputs which correspond to the $2K$ input object queries. 
An MLP is applied to the first $K$ outputs for predicting the coordinates of 3D boxes and their class labels. 
For the rest of the $K$ outputs, we use a different MLP to predict the 2D coordinates of the bounding boxes and their associated classes.
It is worth mentioning that we do not need extra labels for 2D box coordinates, 
since they are obtained by first projecting the labels of 3D box coordinates to the 2D camera plane following \cref{eq:projection}, and then taking the axis-aligned 2D bounding boxes of projected shapes.

The optimization of BrT concerns minimizing a compound loss function that contains two parts: a repression loss for locating bounding boxes, and a classification loss for predicting the class of the associated box.
The regression loss contains two components: $\mathcal{L}_\mathrm{obj}^\mathrm{3D}$ and $\mathcal{L}_\mathrm{obj}^\mathrm{2D}$ for 3D and 2D cases respectively.
Likewise, for classification loss, there are also a 3D component $\mathcal{L}_\mathrm{cls}^\mathrm{3D}$ and a 2D component $\mathcal{L}_\mathrm{cls}^\mathrm{2D}$. As such, the overall loss function is formulated as 
\begin{align}
\label{eq:loss}
\mathcal{L}=\mathcal{L}_\mathrm{obj}^\mathrm{3D}+\alpha_1\mathcal{L}_\mathrm{cls}^\mathrm{3D}+\alpha_2\mathcal{L}_\mathrm{obj}^\mathrm{2D}+\alpha_3\mathcal{L}_\mathrm{cls}^\mathrm{2D},
\end{align}
where $\alpha_1$, $\alpha_2$ and $\alpha_3$ are three parameters weighting relative importance between these losses. In practice, $\mathcal{L}_\mathrm{obj}^\mathrm{3D}$, $\mathcal{L}_\mathrm{obj}^\mathrm{2D}$ or $\mathcal{L}_\mathrm{cls}^\mathrm{3D}$ further consists of two sub-terms. These details will be provided in \cref{sec:datasets_imple}.

\subsection{Transformer building block of BrT}
\label{sec:building-block}

The multi-head attention (MSA) acts as the fundamental building block of the Transformer architecture, which has  three sets of input: query set, key set, and value set.
Given a query set $\{\bm{q}_i\}$ and a common element set $\{\bm{p}_k\}$ of key set and value set, the output feature of MSA of each query element is the aggregation of the values that weighted by linear projections, formulated as:
\begin{equation}
\label{eq:att}
\text{Att}\big(\bm{q}_{i}, \{\bm{p}_{k}\}\big) = \sum_{h=1}^{H} \bm{W}_h  \big( \sum_{k=1}^{K} A^{h}_{i,k} \cdot \bm{V}_h {\bm{p}}_k\big),
\end{equation}
where $h$ indexes over $H$ attention heads; $\bm{W}_h$ and $\bm{V}_h$ are weights for the output projection and the value projection, respectively. 
$A^{h}_{i,k}$ is the attention weight which is computed as
\begin{equation}
\label{eq:att_weight}
A^{h}_{i,k} = \frac{\exp\big[(\bm{Q}_h \bm{q}_{i})^\top (\bm{U}_h \bm{p}_k)\big]}{\sum_{k=1}^{K}{\exp\big[(\bm{Q}_h \bm{q}_{i})^\top (\bm{U}_h \bm{p}_k)\big]}},
\end{equation}
where $\bm{Q}_h$ and $\bm{U}_h$ indicate the query projection weight and the key projection weight, respectively.

As shown in \cref{pic:framework}, during training, the BrT Transformer module prohibits the attentions between $\bm{p}_{\mathrm{pnt}}^{l}$ and $\bm{p}_{\mathrm{pat}}^{l}$.
Therefore, the attention of these tokens are only directly correlated with
\begin{align}
\label{eq:attention-key}
\nonumber
\bm{p}_{\mathrm{pnt},i}^{l+1}&=\text{Att}\big(\bm{p}_{\mathrm{pnt},i}^{l},\{\bm{o}_{\mathrm{pnt},j}^{l}, \; \bm{p}_{\mathrm{pnt},j}^{l}\}\big),\\
\bm{p}_{\mathrm{pat},i}^{l+1}&=\text{Att}\big(\bm{p}_{\mathrm{pat},i}^{l}, \{\bm{o}_{\mathrm{pat},j}^{l}, \; \bm{p}_{\mathrm{pat},j}^{l}\}\big),
\end{align}
where subscript $i$ and $j$ denote token indexes. In practice, calculating the attention scores based on all tokens in the scope yet leaving other tokens unattended can be achieved by applying zero-masking over the obtained attention.

Although we block the attentions between  $\bm{p}_{\mathrm{pnt}}^{l}$ and $\bm{p}_{\mathrm{pat}}^{l}$ in the Transformer modules, 
we choose to explicitly connect them again by adopting the point-to-patch projection, which will be established in \cref{sec:alignment-projection}. 
Instead of allowing global attention in Transformer, 
we believe that the large discrepancy of coordinates essentially
encourages a form of explicit 3D-2D correlation.
We verify this assumption experimentally in \cref{sec:analysis}.

In addition, $\bm{o}_{\mathrm{pnt}}^{l}$ and $\bm{o}_{\mathrm{pat}}^{l}$ have additional functions to further bridge the gap between 3D and 2D coordinates, which will be further detailed in \cref{sec:bridge_by_query}. To this end, the attention of $\bm{o}_{\mathrm{pnt}}^{l}$ and $\bm{o}_{\mathrm{pat}}^{l}$ are correlated  with all tokens,
\begin{align}
\label{eq:attention-query}
\nonumber
\bm{o}_{\mathrm{pnt},i}^{l+1}&=\text{Att}\big(\bm{o}_{\mathrm{pnt},i}^{l},\{\bm{o}_{\mathrm{pnt},j}^{l}, \;\bm{o}_{\mathrm{pat},j}^{l}, \; \bm{p}_{\mathrm{pnt},j}^{l}, \; \bm{p}_{\mathrm{pat},j}^{l}\}\big),\\
\bm{o}_{\mathrm{pat},i}^{l+1}&=\text{Att}\big(\bm{o}_{\mathrm{pat},i}^{l},\{\bm{o}_{\mathrm{pnt},j}^{l}, \;\bm{o}_{\mathrm{pat},j}^{l}, \; \bm{p}_{\mathrm{pnt},j}^{l}, \; \bm{p}_{\mathrm{pat},j}^{l}\}\big).
\end{align}

\subsection{Bridge by conditional object queries}
\label{sec:bridge_by_query}
The Euclidean coordinates of the 3D point cloud may vary dramatically from the camera plane coordinates of 2D image pixels, since they belong to different spaces. 
As a result, it may be hard for Transformer-based model to  learn their relationships even after numerical normalization. 
In this part, we propose to leverage object queries as the bridge for correlating 3D and 2D spaces. 
Specifically, we adopt conditional object queries which are  aware of both the 3D and 2D coordinates.

For Transformer-based object detection models, object queries are observed to probably specialize on certain areas and box sizes during the training process, even they are generated by random initialization~\cite{carion2020end}. 
Inspired by this, we assume that hidden features of object queries {w.r.t} points and images could be potentially aligned inside the Transformer.  
Hence instead of using randomly generated object queries, we adopt conditional object queries to boost the prediction learning process based on the object query alignment of points and images. 

To align object queries of points and image patches, we first sample $K$ points as  proposals from the $N_{\mathrm{pnt}}$ points with kNN search, and denote the 3D coordinates and features of these $K$ points as $\bm{k}_{\mathrm{pnt}}\in\mathbb{R}^{K\times 3}$ and $\bm{f}_{\mathrm{pnt}}\in\mathbb{R}^{K\times F}$, respectively. We then learn the 3D coordinates of proposals, denoted as $\bm{k}'_{\mathrm{pnt}}\in\mathbb{R}^{K\times 3}$, by adding $\bm{k}_{\mathrm{pnt}}$ with additional learned biases based on $\bm{f}_{\mathrm{pnt}}$. The object queries for points $\bm{o}_{\mathrm{pnt}}^{1}$ are conditioned on the $\bm{k}'_{\mathrm{pnt}}$. Formally, there is
\begin{align}
\label{eq:k-point}
\bm{k}'_{\mathrm{pnt}}&=\bm{k}_{\mathrm{pnt}}+\text{MLP}(\bm{f}_{\mathrm{pnt}}),\\
\label{eq:o-point1}
\bm{o}_{\mathrm{pnt}}^{1}&=\text{MLP}(\bm{k}'_{\mathrm{pnt}})+\bm{\mathrm{PE}},
\end{align}
where $\bm{\mathrm{PE}}\in\mathbb{R}^{K\times D}$ is the randomly initialized positional embeddings.

Regarding the object queries of image patches $\bm{o}_{\mathrm{pat}}^{1}$, we project $\bm{k}'_{\mathrm{pnt}}$ to the corresponding image and obtain the 2D coordinates of projected pixels, denoted as $\bm{\mathrm{proj}}(\bm{k}'_{\mathrm{pnt}})\in\mathbb{R}^{K\times 2}$ where $\bm{\mathrm{proj}}$ indicates projecting points to the image according to the per-point projection in \cref{eq:projection}. The object queries for image patches are conditionally obtained by 
\begin{align}
\label{eq:o-patch1}
\bm{o}_{\mathrm{pat}}^{1}=\text{MLP}\big(\bm{\mathrm{proj}}(\bm{k}'_{\mathrm{pnt}})\big)+\bm{\mathrm{PE}},
\end{align}
where $\bm{\mathrm{PE}}$ is the same positional embeddings as in \cref{eq:o-point1}. Sharing positional embeddings intuitively tells the  Transformer that both object queries $\bm{o}_{\mathrm{pnt}}^{1}$ and $\bm{o}_{\mathrm{pat}}^{1}$ are aligned.  

Up till here, we improve the designs of object queries to bridge the representation spaces of 3D point clouds and 2D images. 
Such design is described by \cref{eq:attention-query}, \cref{eq:o-point1}, and \cref{eq:o-patch1}, which actually embody two perspectives: the attentive connections and the alignment of 3D and 2D object queries using a shared $\bm{\mathrm{PE}}$. 
Ablation studies in \cref{sec:analysis} verify the effectiveness of  both components we propose.

\subsection{Bridge by point-to-patch projection}
\label{sec:alignment-projection}
Apart from correlating point tokens $\bm{p}_{\mathrm{pnt}}^{l}$ and  patch tokens $\bm{p}_{\mathrm{pat}}^{l}$ with object queries,
we further strengthen their relations by adding intrinsic point-to-patch projection. 
Denoting the 3D coordinates of $N_{\mathrm{pnt}}$ sampled points as $\bm{n}_{\mathrm{pnt}}\in\mathbb{R}^{N_{\mathrm{pnt}}\times 3}$, we project $\bm{n}_{\mathrm{pnt}}$ to the corresponding camera plane and obtain $N_{\mathrm{pnt}}$ 2D pixel coordinates that are denoted with $\bm{\mathrm{proj}}(\bm{n}_{\mathrm{pnt}})\in\mathbb{R}^{N_{\mathrm{pnt}}\times 2}$.
With $\bm{\mathrm{proj}}$ already defined in \cref{eq:o-patch1}, 
we can conveniently let $\bm{u}_n$ and $\bm{v}_n$ be the $x$-axis value and the $y$-axis value respectively of the $n$-th element of $\bm{\mathrm{proj}}(\bm{n}_{\mathrm{pnt}})$, where $n=1,2,\cdots,N_{\mathrm{pnt}}$. 
If $\bm{u}_n$ and $\bm{v}_n$ respectively satisfy restrictions $1\le\bm{u}_n\le H$ and $1\le\bm{v}_n\le W$, 
then such a 2D coordinate could reside in the input image of size $H\times W\times 3$. 
Rounding $\bm{u}_n$ and $\bm{v}_n$ to the nearest integers obtains the valid coordinates indicating certain image pixels. It is now easy to obtain the corresponding image patch index with
\begin{align}
\label{eq:get-patch}
\bm{p}_n=
\big\lfloor{\lfloor \bm{v}_n  \rfloor}/{S}\big\rfloor\times \big\lfloor{W}/{S}\big\rfloor+\big\lfloor{\lfloor \bm{u}_n \rfloor}/{S}\big\rfloor,
\end{align}
where $\lfloor\cdot\rfloor$ is the rounding operator; $\bm{p}_n\in\{1,2,\ldots,N_{\mathrm{pat}}\}$ is the corresponding patch index for the $n$-th point; $S$ denotes the image patch size.
Our point-to-patch projection aggregates features for both points and image patches by
\begin{align}
\label{eq:patch-to-token-projection}
\bm{p}_{\mathrm{pnt},n}^{l}=\bm{p}_{\mathrm{pnt},n}^{l}+\text{MLP}\big(\bm{p}_{\mathrm{pat},\bm{p}_n}^{l}\big),
\end{align}
where subscripts $n$ and $\bm{p}_n$ indicate indexes of the token features $\bm{p}_{\mathrm{pnt}}^{l}$ and $\bm{p}_{\mathrm{pat}}^{l}$, respectively.

\subsection{Extend to multiple-view scenarios}
\label{sec:multiview}
It is challenging to directly extend current point-only or point-image methods to  detect from the point cloud and \textbf{multi-view} images, which is nevertheless  a common real-life situation of data organization. 
For example, \cite{qi2020imvotenet} avoided using the ScanNetV2 dataset which contains rich multi-view images, probably due to the difficulty in  combining  interactions of point  cloud and each view.

Fortunately, with few bells and whistles, our proposed BrT has a natural advantage in combining the point cloud with multi-view images, where both point-image interactions and the interactions of  multi-view images can be utilized for further improved performance. As shown in \cref{pic:framework}, when there are different views of input images for one single scene, we first concatenate these images along the width-side and obtain a wide image. The following processes are the same with the single-view condition. Since for the multi-view images, each view usually contains fewer objects, we expect the number of object queries $K$ can still handle all the objects. Our current design mainly aims to bridge between each view and the point cloud, yet it does not exploit the relations among different views with explicit projections, which is left to be our future work.

\section{Experiments}
\label{sec:experiments}
Our experiments are conducted on the challenging SUN RGB-D~\cite{song2015sun} and ScanNetV2~\cite{DBLP:conf/cvpr/DaiCSHFN17} datasets. We first detail the settings for datasets and implementations in \cref{sec:datasets_imple}. Then, we quantitatively compare our BrT with the state-of-the-art methods in \cref{sec:comparion_sota}; and we present and discuss the qualitative results in \cref{sec:qualitative}. Finally, we perform analytical experiments in \cref{sec:analysis} to verify the advantage of each component in BrT. More details of network architectures and visualizations are provided in our Appendix.

\subsection{Datasets and implementation details}
\label{sec:datasets_imple}
\textbf{Datasets.} SUN RGB-D~\cite{song2015sun} is a single-view RGB-D dataset for 3D scene understanding. It consists of 10,335 RGB-D images annotated with amodal oriented 3D bounding boxes for $37$ object categories, alongside corresponding camera poses. The training and validation splits are composed of 5,285 and 5,050 frames respectively. We convert  depth images to point clouds using the provided camera parameters, and adopt a standard evaluation protocol to report performance on the $10$ most common categories~\cite{qi2019deep,qi2020imvotenet,liu2021group}.

\begin{table*}[t]
	\centering
	\tablestyle{4pt}{0.98}
	\resizebox{1\linewidth}{!}{
		\begin{tabular}{l|c|p{0.8cm}<{\centering}p{0.8cm}<{\centering}p{0.8cm}<{\centering}p{0.8cm}<{\centering}p{0.8cm}<{\centering}p{0.8cm}<{\centering}p{0.8cm}<{\centering}p{0.8cm}<{\centering}p{0.8cm}<{\centering}p{0.8cm}<{\centering}|cc}
			\shline
			\rowcolor{gray!12}
			\hskip0.02in Method & RGB & bathtub & bed & bookshf & chair & desk & dresser & nightstd & sofa & table & toilet & mAP@$0.25$ & mAP@$0.5$ \\
\shline
VoteNet~\cite{qi2019deep} & \xmark & 75.5 &85.6&31.9&77.4&24.8&27.9&58.6& 67.4&51.1&90.5&59.1 &35.8 \\
MLCVNet~\cite{xie2020mlcvnet} & \xmark &  79.2 & 85.8 & 31.9 & 75.8 & 26.5 & 31.3 & 61.5 & 66.3 & 50.4 & 89.1 & 59.8&- \\
HGNet~\cite{chen2020hierarchical} & \xmark & 78.0 &84.5&  {35.7}& 75.2&  {34.3} & 37.6& 61.7& 65.7& 51.6& {91.1}& 61.6&- \\
H3DNet~\cite{zhang2020h3dnet}& \xmark & 73.8 &85.6 &31.0& 76.7 &29.6& 33.4& 65.5& 66.5& 50.8& 88.2& 60.1&39.0 \\
\hline
Group-Free~\cite{liu2021group}& \xmark& {80.0} & {87.8} & 32.5  & {79.4} &32.6 & {36.0} & {66.7}  & {70.0}  & {53.8}  &{91.1}  & {63.0}&45.2\\  
\rowcolor{gray!8}
$\;\;$+$3$-dim RGB& \cmark& {77.1} & {87.2} & 31.2  & {76.5} &30.8 & {36.2} & {66.3}  & {68.1}  & {53.0}  &{90.7}  & {61.7} &42.0 \\  
\rowcolor{gray!8}
$\;\;$+Faster R-CNN& \cmark& {78.0} & {87.4} & 34.3  & {77.2} &32.8 & {36.5} & {67.0}  & {68.8}  & {53.2}  &{91.6}  & {62.7} &44.2 \\  
\rowcolor{gray!8}
$\;\;$+YOLOS& \cmark& {80.6} & {87.5} & 35.0  & {78.5} &32.2 & {37.3} & {66.7}  & {69.3}  & {54.4}  &{92.1}  & {63.4} &45.7\\  
\hline
DSS~\cite{song2016deep} &\cmark & 44.2 & 78.8 & 11.9 & 61.2 & 20.5 & 6.4 & 15.4 & 53.5 & 50.3 & 78.9 & 42.1  &-  \\
2D-driven~\cite{lahoud20172d} & \cmark & 43.5 & 64.5 & 31.4 & 48.3 & 27.9 & 25.9 & 41.9 & 50.4 & 37.0 & 80.4 & 45.1&-  \\
PointFusion~\cite{xu2018pointfusion} & \cmark & 37.3 & 68.6 & 37.7 & 55.1 & 17.2 & 23.9 & 32.3 & 53.8 & 31.0 & 83.8 & 45.4&- \\ 
F-PointNet~\cite{qi2018frustum} &  \cmark& 43.3 & 81.1 & 33.3 & 64.2 & 24.7 & 32.0 & 58.1 & 61.1 & 51.1 &{90.9} & 54.0&- \\ 
imVoteNet~\cite{qi2020imvotenet} & \cmark &{75.9} &{87.6} &\textbf{41.3} & 76.7 & 28.7 &\textbf{41.4} &\textbf{69.9} &{70.7} & 51.1 & 90.5 &{63.4}&- \\
\hline
Our BrT &  \cmark & \textbf{82.8} &\textbf{88.0}&40.5&\textbf{79.7}&\textbf{33.4}&{40.6}&67.4&\textbf{71.1}&\textbf{55.7}&\textbf{93.5}&\textbf{65.4}&\textbf{48.1}\\
			\shline
		\end{tabular}
	}\vskip-0.035in
\caption{\textbf{3D object detection results on the SUN RGB-D validation set.} Evaluation metrics include the average precisions with 3D IoU threshold 0.25 (mAP@0.25) and threshold 0.5 (mAP@0.5), respectively, as proposed by~\cite{song2015sun}. Single-class metric adopts mAP@0.25 for evaluation. All listed methods adopt the geometric information (depth or point cloud), and a part of them additionally utilize RGB as input.}
\label{table:sunrgbd}
\vskip-0.1in
\end{table*}

ScanNetV2~\cite{DBLP:conf/cvpr/DaiCSHFN17} is a richly annotated dataset of 3D reconstructed meshes of indoor scenes. It contains 1,513 scans covering more than 700 unique indoor scenes, out of which 1,201 scans belong to the training split, and the rest 312 scans comprise the validation subset. ScanNetV2 contains over 2.5 million images with camera poses, and their corresponding reconstructed point clouds with 3D semantic annotation for 18 object categories. Compared to single-view scans in SUN RGB-D, scenes in ScanNetV2 are more complete and cover larger areas with multiple views. In our experiments, we adopt the sample dataset from ScanNetV2 containing 25,000 frames ($\mathrm{scannet\_frames\_25k}$) which are sampled with 100 interval frames from the whole dataset. 

\textbf{Implementation details.} Following the common successful practice in~\cite{qi2019deep,liu2021group}, we adopt PointNet++~\cite{DBLP:conf/nips/QiYSG17} as the point cloud backbone. The backbone has four set abstraction layers where the input point cloud is sub-sampled to 2,048, 1,024, 512, and 256 points with the increasing receptive radius of 0.2, 0.4, 0.8, and 1.2, respectively. There are two feature propagation layers which successively up-sample the points to 512 and 1,024, i.e. $N_{\mathrm{pnt}}$=1,024. 

The point cloud is augmented  following \cite{qi2019deep} that employs random flipping, random rotation between $-5^{\circ}$ and $5^{\circ}$, and random scaling with a factor from 0.9$\times$ to 1.1$\times$.
We use 20k and 50k points as input for each point cloud on SUN RGB-D and ScanNetV2 datasets, respectively.  For ScanNetV2, we use depths to filter out the projected 3D points which should be occluded, but visible due to the sparsity of the point cloud. Since ScanNetV2 does not provide the oriented bounding box annotation, we predict axis-aligned bounding boxes without the rotation angle, as in~\cite{qi2019deep,liu2021group}.

For default experiments, model parameters are initialized to weights of ViT-S/16, pre-trained on ImageNet-$1k$. Yet, the MLP heads for classification and bounding box regression and  object queries are generated by \cref{eq:o-point1} and \cref{eq:o-patch1}. We set the hidden size to 384, the layer depth to 12, the patch size to 16, and the number of attention heads to 6. Images are resized to $530\times 730$ , with $N_{\mathrm{pat}}$=$\lfloor530/16\rfloor\times\lfloor730/16\rfloor$=1,485. Besides, there is $K$=256.

For the SUN RGB-D dataset, we include an additional orientation prediction branch to predict the orientation of the 3D box, which additionally includes a classification task and an offset regression task with loss weights of 0.1 and 0.04, respectively following~\cite{liu2021group}.

\begin{figure*}[t!]
\centering\vskip-0.05in
\includegraphics[scale=0.145]{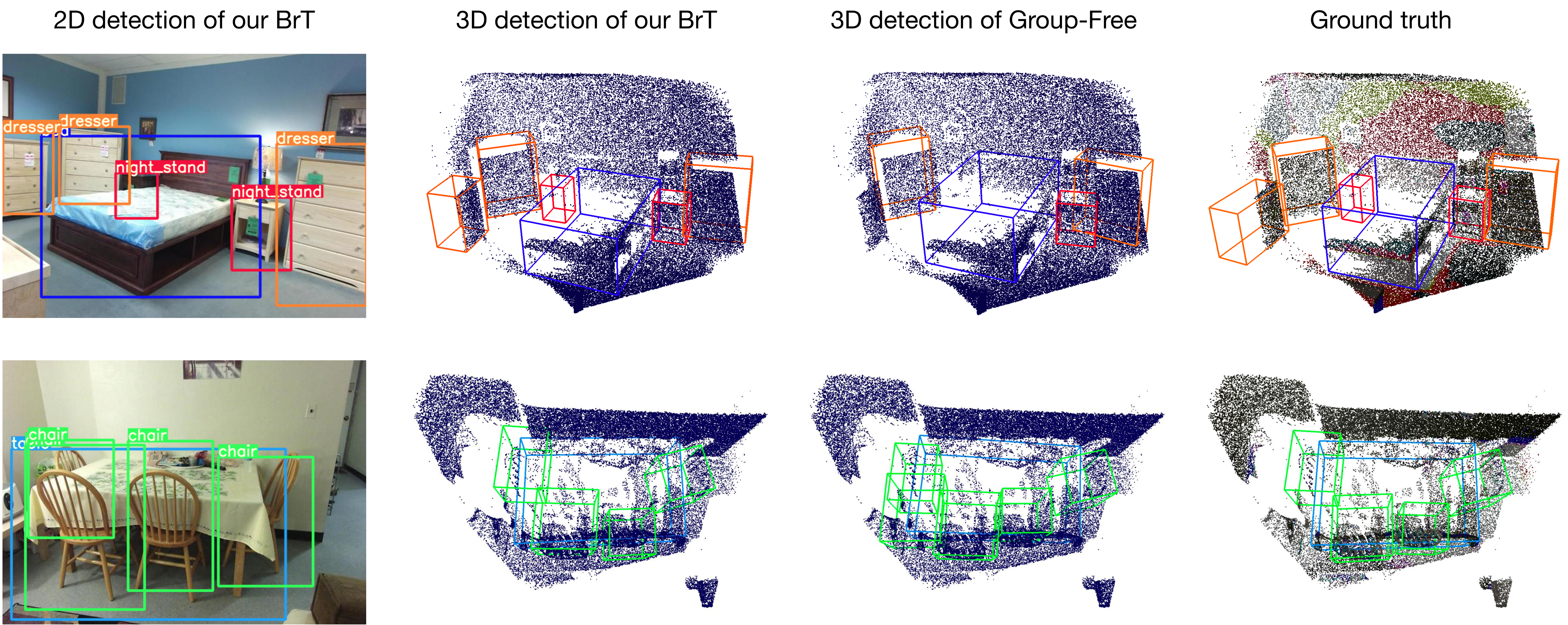}
\caption[]{Visualization comparisons to show the advantage of using image information. We compare our BrT with Group-Free~\cite{liu2021group} which adopts the point cloud as input. First row: Our BrT detects nightstand even the points are very sparse with the help of image detection. Second row: Group-Free is affected by the noises and  detects a  false positive chair, while BrT is relatively robust.}
\label{pic:compare-gf}
\end{figure*}

We train BrT with the AdamW~\cite{DBLP:journals/corr/abs-1711-05101} optimizer ($\beta_1$=0.9, $\beta_2$=0.999) with $600$ epochs. The learning rate is initialized to $0.004$ and decayed by 0.1$\times$ at the $420$-th epoch, the $480$-th epoch, and the $540$-th epoch. We set the loss weights in \cref{eq:loss} to $\alpha_1=0.2$, $\alpha_2=0.5$ and $\alpha_3=0.1$. Following~\cite{qi2019deep,liu2021group}, $\mathcal{L}_\mathrm{obj}^\mathrm{3D}$ consists of a center offset sub-loss and a size offset sub-loss with weights $1.0$ and $0.1$, respectively; $\mathcal{L}_\mathrm{cls}^\mathrm{3D}$ consists of an object classification sub-loss and a size classification sub-loss with equal weights $1.0$. In addition, following~\cite{fang2021you}, $\mathcal{L}_\mathrm{obj}^\mathrm{2D}$ consists of a center offset sub-loss and a GIoU sub-loss with weights $1.0$ and $2.0$, respectively.

\subsection{Comparison with state-of-the-art methods}
\label{sec:comparion_sota}

In this section, we compare our BrT with state-of-the-art methods (including using RGB-depth, RGB-point cloud, or point cloud only) for 3D object detection on both SUN RGB-D and ScanNetV2 datasets.

\textbf{Results on SUN RGB-D.}
In Table \ref{table:sunrgbd}, we provide detailed per-class 3D object detection results on SUN RGB-D. We observe that our BrT achieves new records which are  remarkably superior to previous methods in terms of mAP@$0.25$ and mAP@$0.5$. Specifically,  BrT surpasses Group-Free~\cite{liu2021group}, which is a Transformer-based model with only point clouds as input, by $2.3$\% (mAP@$0.25$) and $2.9$\% (mAP@$0.5$); and surpasses imVoteNet~\cite{qi2020imvotenet}, which is the current best CNN-based model also using RGB, by $1.9$\% (mAP@$0.25$). Note that Group-Free adopts a multi-stage ensemble over all Transformer stages to boost the performance, while our BrT uses one output for evaluation.

Since Group-Free~\cite{liu2021group} achieves the best performance among the methods that only use the geometric information, we experiment with three additional Group-Free variants: (1) ``+3-dim RGB'' directly appends the the $3$-dimensional RGB values to the point cloud features (of the seed points sampled by PointNet++); (2) ``+Faster R-CNN'' adopts a pre-trained Faster R-CNN~\cite{ren2015faster} (same model used in~\cite{qi2020imvotenet}), a CNN-based 2D detector, to extract region features and concatenate them to the seed points inside that 2D box frustum; (3) ``+YOLOS'' adopts a pre-trained YOLOS~\cite{fang2021you} (with DeiT-S~\cite{pmlr-v139-touvron21a} model), a Transformer-based 2D detector, to extract image patch features and project them by our bridging method in \cref{sec:alignment-projection}. 

Results of these three variants (with RGB) are also provided in Table \ref{table:sunrgbd}, where only ``+YOLOS'' achieved marginal grain. This result indicates that intuitive integration of RGB information is difficult to boost the performance. Besides, simply appending 3-dimensional RGB to the point features even impacts the performance, which we conjecture is owing to the discrepancy of 2D/3D representations. By comparison, our BrT is notably better than these three variants.

\begin{table}[t]
	\centering
	\tablestyle{1pt}{1}
	\resizebox{1\linewidth}{!}{
		\begin{tabular}{l|c|c|cc}
			\shline
			\rowcolor{gray!12}
			\hskip0.02in Method
			& Backbone
			& RGB
			& mAP@$0.25$
			& mAP@$0.5$\\
			\hline
			VoteNet~\cite{qi2019deep} & PointNet++ & \xmark&62.9 & 39.9 \\
			MLCVNet~\cite{xie2020mlcvnet} & PointNet++ & \xmark& 64.5 & 41.4 \\
			H3DNet~\cite{zhang2020h3dnet} & PointNet++ & \xmark& 64.4 & 43.4 \\
			H3DNet~\cite{zhang2020h3dnet} & 4$\times$PointNet++ & \xmark& 67.2 & 48.1 \\
			HGNet~\cite{chen2020hierarchical}&GU-net & \xmark& 61.3 & 34.4 \\
			GSDN~\cite{gwak2020generative} & MinkNet & \xmark& 62.8 & 34.8 \\
			3D-MPA~\cite{engelmann20203d} & MinkNet& \xmark& 64.2 & 49.2 \\

			\hline
			\makecell[l]{Group-Free~\cite{liu2021group} }(12-L)& PointNet++ & \xmark & 67.3 & 48.9  \\
			\makecell[l]{Group-Free~\cite{liu2021group}} (24-L)& PointNet++w2$\times$ &\xmark & {69.1}  & {52.8} \\
\rowcolor{gray!8}
			$\;\;$+$3$-dim RGB& PointNet++w2$\times$ & \cmark & {67.8}  & {51.0}\\
\rowcolor{gray!8}
			$\;\;$+Faster R-CNN& PointNet++w2$\times$ & \cmark & {68.7}  & {52.2} \\
\rowcolor{gray!8}
			$\;\;$+YOLOS& PointNet++w2$\times$ & \cmark & {69.2}  & {52.6} \\
\hline
			\makecell[l]{{Our BrT} (ViT-S/16) } &PointNet++ & \cmark & {69.7}  &{53.0} \\
			\makecell[l]{{Our BrT} (ViT-B/16)} &PointNet++w2$\times$ & \cmark & \textbf{71.3}  & \textbf{55.2}  \\
			\shline
		\end{tabular}
	}
\vskip-0.035in
\caption{\textbf{3D object detection results on the ScanNetV2 validation set.} All listed methods adopt the geometric information from point clouds. L denotes the number of self/cross-attention layers. PointNet++w2$\times$ expands the backbone width by 2 times.}
\vskip-0.035in
\label{table:scannet}
\end{table}

\textbf{Results on ScanNetV2.} Table \ref{table:scannet} provides performance comparison on ScanNetv2. Similarly, we also conduct three additional experiments including ``+3-dim RGB'', ``+Faster R-CNN'', and ``+YOLOS'' as competitive baselines based on the state-of-the-art method Group-Free (24-L)~\cite{liu2021group}. We observe that these three experiments also fail to bring noticeable improvements. We experiment with two architectures for our BrT which adopt the structural designs of ViT-S/16 and ViT-B/16, respectively. The light architecture BrT (ViT-S/16) already surpasses all compared methods.  BrT (ViT-B/16) further obtains an additional gain of  1.6 mAP.

\subsection{Qualitative results and discussion}
\label{sec:qualitative}
In \cref{pic:compare-gf}, we compare the state-of-the-art method Group-Free~\cite{liu2021group} (with only point cloud input) with our BrT (with additionally image input). We observe that with the help of image clues, our BrT identifies the nightstand which is partially behind the bed acquiring few points, while Group-Free fails to detect it. In addition, in the second case, Group-Free detects a false positive chair due to the point noises, and our BrT seems to be robust to point noises thanks to the image information. We provide more visualizations on ScanNetv2 with multi-view images in our Appendix.

\begin{figure}[t!]
\centering\vskip-0.05in
\includegraphics[scale=0.145]{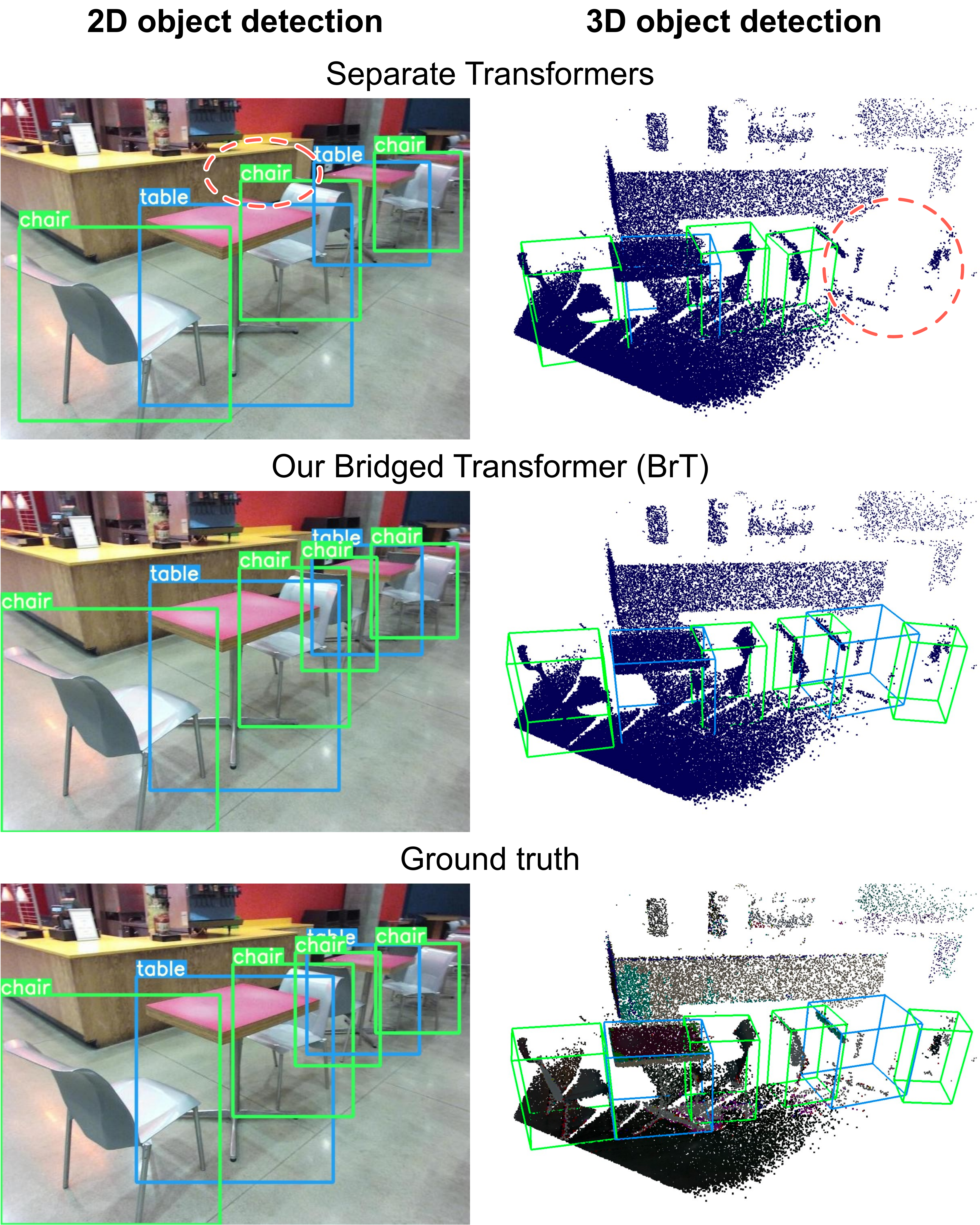}
\caption[]{Visualization comparisons of using separate Transformers and our BrT. In the first row, some less obvious objects are missing (one of the occluded 2D chairs, and also the table/chair with sparse 3D points) which are highlighted with red circles. Our BrT successfully captures these difficult objects.}
\label{pic:wo-bridge}\vskip-0.05in\vskip-0.05in
\end{figure}

\begin{table}[t]
	\centering
	\tablestyle{2pt}{1}
	\resizebox{1\linewidth}{!}{
		\begin{tabular}{c|c|c|cc}
			\shline
			\rowcolor{gray!12}
			\makecell[c]{Conditional\\object queries}  
			& \makecell[c]{Shared PEs of\\object queries} 
			& \makecell[c]{point-to-patch\\projection}   
			& mAP@$0.25$
			& mAP@$0.5$\\
			\hline
\xmark & \xmark &  \xmark & {62.6}  &{44.3} \\
  \cmark & \xmark&  \xmark & {62.9}  &{44.7} \\
  \hline
    \xmark & \cmark &  \xmark & {62.4}  &{44.0} \\
    \cmark & \cmark &  \xmark & {64.0}  &{46.3} \\
    \hline
 \xmark&   \xmark&  \cmark & {64.2}  &{46.6} \\
  \cmark & \cmark & \cmark & \textbf{65.4}  &\textbf{48.1}\\
			\shline
		\end{tabular}
	}
\vskip-0.035in
\caption{Ablation studies of our bridging techniques, including results when using/not using conditional object queries, shared PEs, and the point-to-patch projection proposed in \cref{sec:bridge_by_query} and \cref{sec:alignment-projection}. Experiments are performed on  SUN RGB-D.}
\label{table:ablation-bridge}
\end{table}

\begin{table}[t]
	\centering
	\tablestyle{0.5pt}{0.98}
	\resizebox{1\linewidth}{!}{
		\begin{tabular}{c|c|c|cc}
			\shline
			\rowcolor{gray!12}
			\makecell[c]{Connects. between\\ $\bm{o}_{\mathrm{pnt}}^l$ and $\bm{p}_{\mathrm{pat}}^l$}  
			& \makecell[c]{Connects. between\\ $\bm{o}_{\mathrm{pat}}^l$ and $\bm{p}_{\mathrm{pnt}}^l$}  
			& \makecell[c]{Connects. between\\ $\bm{p}_{\mathrm{pnt}}^l$ and $\bm{p}_{\mathrm{pat}}^l$}     
			& mAP@$0.25$
			& mAP@$0.5$\\
			\hline
\xmark & \xmark &  \xmark& {62.4}  &{44.3} \\
\cmark & \xmark&  \xmark & {64.5}  &{47.0} \\
\xmark & \cmark &  \xmark& {62.8}  &{45.0} \\
\hline
\cmark& \cmark &  \cmark & {63.9}  &{46.4} \\
    \cmark & \cmark & \xmark & \textbf{65.4}  &\textbf{48.1}\\
			\shline
		\end{tabular}
	}
\vskip-0.035in
\caption{Comparisons of different attention connections (connects.) in BrT with our default version (last row) proposed in \cref{sec:building-block}. Experiments are performed on  SUN RGB-D.}
\label{table:ablation-connect}
\end{table}

\begin{table}[t]
	\centering
	\tablestyle{7pt}{1.2}
	\resizebox{1\linewidth}{!}{
		\begin{tabular}{>{\columncolor{gray!12} }c|p{1.4cm}<{\centering}|p{1.4cm}<{\centering}|p{1.4cm}<{\centering}}
			\shline
			
			\makecell[c]{Pre-trained models}  
			& No pre-train & ViT-S
			& YOLOS-S
			\\
			\hline
			mAP results& 62.8 (44.5)&\textbf{65.4 (48.1)}& 63.4 (45.3) \\
			\shline
		\end{tabular}
	}
\vskip-0.035in
\caption{Results of mAP@$0.25$ and mAP@$0.5$ (in brackets) when using different pre-training strategies. Experiments are performed on SUN RGB-D.}
\label{table:pretrain}\vskip-0.05in
\end{table}

To highlight the advantage of our bridging techniques proposed in \cref{sec:bridge_by_query} and \cref{sec:alignment-projection}. In \cref{pic:wo-bridge}, we illustrate visualization comparisons without or with 3D\&2D bridging. Specifically, in the first row, we adopt two separate Transformers to learn 2D objects and 3D objects respectively, where the learning processes of both Transformers are isolated without any bridging techniques. Highlighted with  red circles, the result of 2D detection fails to identify the two overlapped chairs, and instead, predicts them as one single chair. In addition, its 3D detection result ignores a table and a chair due to the sparsity of points. This is a common bottleneck of 3D object detection, as points are usually unevenly distributed resulting in many sparse regions. Conversely, our BrT in the second row successfully predicts both overlapped chairs correctly in 2D detection, which is probably a consequence of the disambiguation learned by BrT from the 3D space. 
In addition, with the help of the image, BrT also captures all tables and chairs in 3D detection, even in the regions where points are extremely sparse.

\subsection{Ablation analysis}
\label{sec:analysis} 
In this section, we perform ablation studies to isolate the benefit of each important component in BrT.

\textbf{Bridging technique.} In \cref{sec:bridge_by_query} and \cref{sec:alignment-projection}, we propose to use conditional object queries, shared PEs of object queries, and the point-to-patch projection. We provide detailed comparison results in Table \ref{table:ablation-bridge} to verify the necessity of each component. We observe that the combination of conditional object queries and shared PEs are mandatory to ensure a significant performance uplift, since only using one of them does not lead to a noticeable improvement. In addition, the point-to-patch projection itself further boosts the performance. These results indicate that all of the three components are necessary.

\textbf{Attention connection.} As proposed in \cref{sec:building-block}, for each layer $l$, we block the direct attention connections between the point tokens $\bm{p}_{\mathrm{pnt}}^l$ and image patch tokens $\bm{p}_{\mathrm{pat}}^l$, but we adopt globally attentive object queries $\bm{o}_{\mathrm{pnt}}^l$ and $\bm{o}_{\mathrm{pat}}^l$ to correlate points and image patches. To demonstrate the advantage of this design, in Table~\ref{table:ablation-connect}, we compare different kinds of connections. We observe that using globally attentive  object queries indeed help improve the performance, and our default connection setting (last row) achieves the best performance.  Connections between $\bm{p}_{\mathrm{pnt}}^l$ and $\bm{p}_{\mathrm{pat}}^l$ lead to performance drops while increase computation costs, which are the reasons that we block these connections.

\textbf{Pre-training strategy.} As described in \cref{sec:datasets_imple}, our BrT is initialized with ViT-S parameters pre-trained on ImageNet-$1k$. To verify the transferability of using such pre-training strategy, in Table~\ref{table:pretrain}, we first compare our results with training from scratch. By comparison, using ViT-S pre-training achieves 2.6 higher mAP@0.25 than training from scratch, which demonstrates the good transfer learning capability of our model, encouraging applying our architecture on larger datasets. In addition, we also try to initialize BrT with off-the-shelf parameters of YOLOS, which is first pre-trained on ImageNet-$1k$ and then finetuned on the COCO dataset~\cite{lin2014microsoft}. Yet we do not observe further improvements when using COCO finetuned parameters, and we speculate that parameters finetuned on COCO may be easy to overfit the SUN RGB-D dataset.

\section{Conclusion}
This work presents BrT for 3D object detection that is able to  exploit the point clouds and images effectively. Due to the large discrepancy between the representations of 2D and 3D spaces, point tokens and image patch tokens in BrT are correlated with globally attentive object queries, instead of directly connected. In addition, we propose to bridge the 2D and 3D learning processes from two perspectives including leveraging conditional object quires and the point-to-patch projection. Our BrT surpasses state-of-the-art methods on SUN RGB-D and ScanNetV2 datasets, which also demonstrate the scalability of BrT in multi-view scenarios.

\section*{Acknowledgement}
This work is funded by Major Project of the New Generation of Artificial Intelligence (No. 2018AAA0102900), the Sino-German  Collaborative Research Project Crossmodal Learning (NSFC  62061136001/DFG TRR169) and Major Science and Technology Innovation 2030 ``New Generation Artificial Intelligence'' key project (No. 2021ZD0111700). We  acknowledge the support of MindSpore, CANN  and Ascend AI Processor used for this research.

\begin{figure*}[t!]
\centering
\hskip0.1in
\includegraphics[scale=0.575]{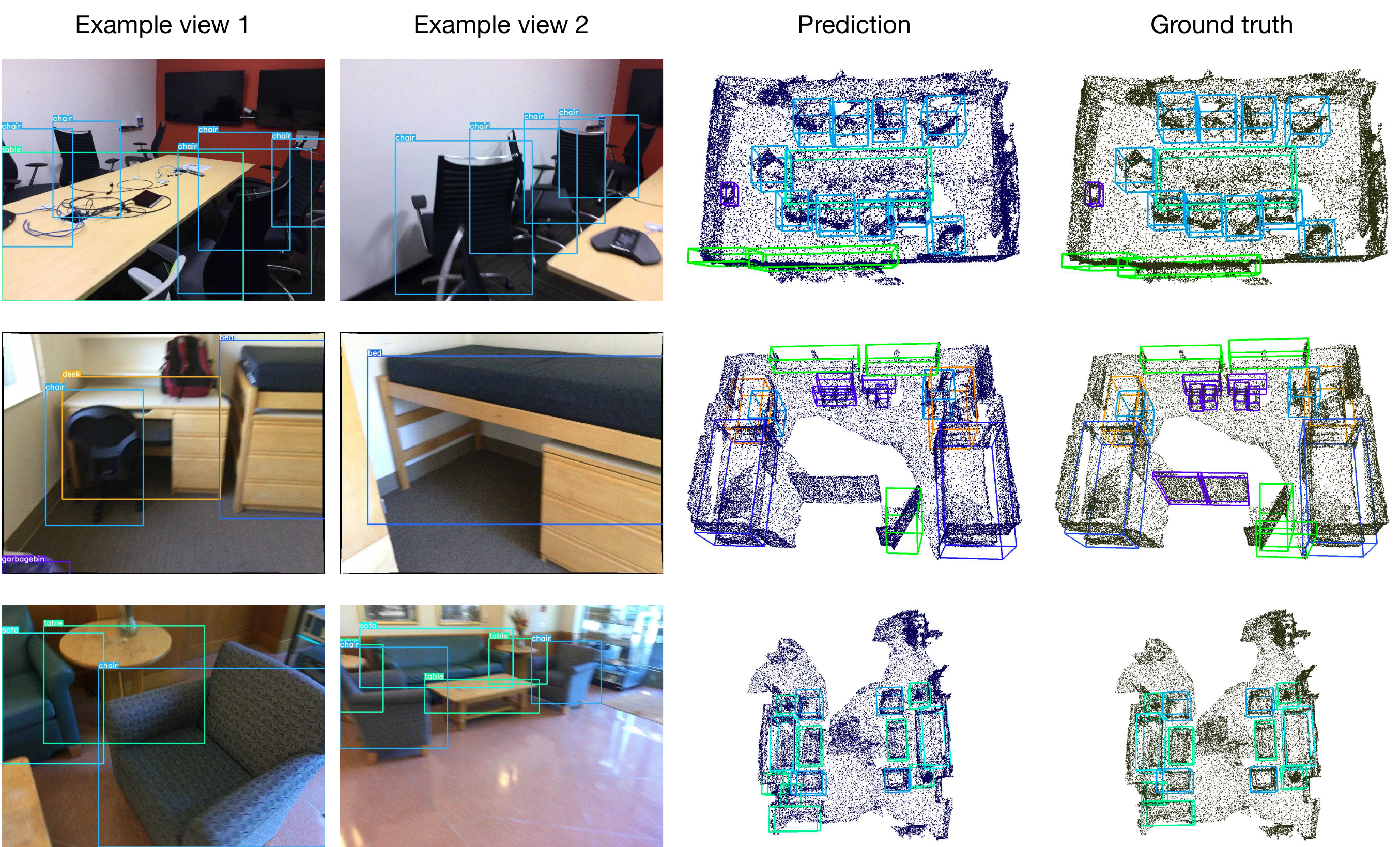}
\caption[]{Additional visualizations on the ScanNetV2 dataset based on multi-view images and the point cloud as input. Two predicted image views are provided as examples.}
\label{pic:additional_visual}
\end{figure*}

\section*{\LARGE Appendix}
\appendix

\section{Implementation Details}
\label{sec:implementation}
 On both datasets, our implementation for the point cloud mostly follows implementation settings in~\cite{qi2019deep,liu2021group}. This part provides additional implementation details. For SUN RGB-D, we use 20k points as input for each point cloud, and these points are randomly sampled from the depth image. For ScanNetV2, we adopt 50k points as input, which are randomly sampled  from the scanned point cloud. We augment sampled points by flipping along the YZ plane with the probability of $50\%$. Note that the 3D box labels are also flipped or rotated following the augmentation of points. Besides, to preserve the lifting relations after the augmentation of points, we define an augmented matrix as
\begin{align}
\label{eq:projection}
\bm{A} =\small
\left[ \begin{array}{ccc}
\cos(\theta)\cdot\mathbb{I}_\mathrm{flip}\cdot s & \sin(\theta)\cdot\mathbb{I}_\mathrm{flip} & 0 \\
-\sin(\theta) &\cos(\theta)\cdot s & 0 \\
0 & 0 & s
\end{array} \right],
\end{align}
where $\theta$  is the rotation angle of points along the Z-axis; $\mathbb{I}_\mathrm{flip}$ denotes the indicator function which is $1$ if points are flipped along the YZ plane, otherwise $-1$; $s$ is the overall scaling ratio of the point cloud.

\begin{figure}[t!]
\centering
\vskip0.13in
\includegraphics[scale=0.085]{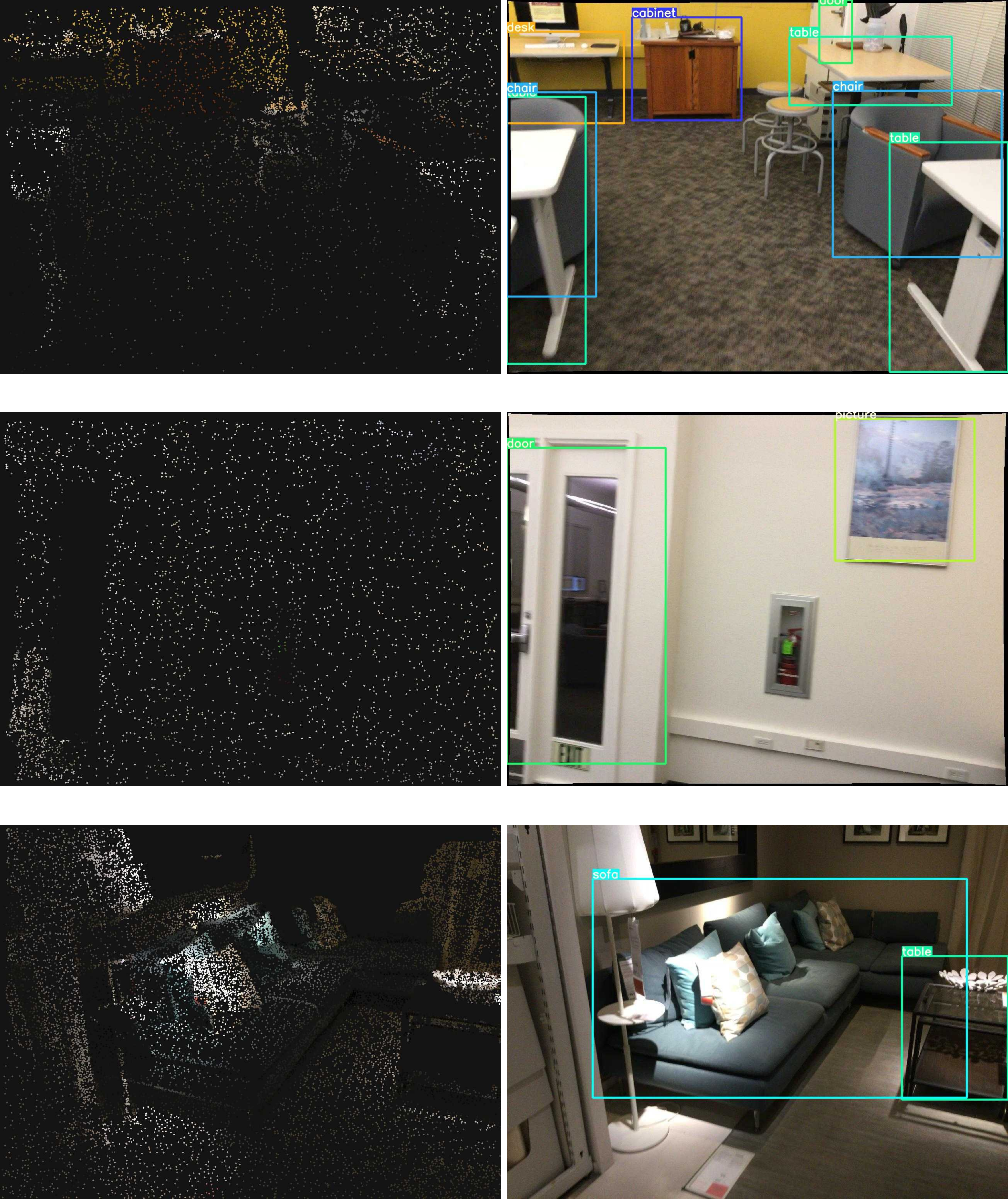}
\caption[]{\textbf{Left:} Projecting points  to multi-view images. \textbf{Right:} Projecting 3D bounding boxes as eight points to multi-view images. Images are collected from the ScanNetV2 dataset.}
\vskip-0.3in
\label{pic:points_boxes}
\end{figure}

\begin{figure*}[t!]
\centering
\hskip0.1in
\includegraphics[scale=0.29]{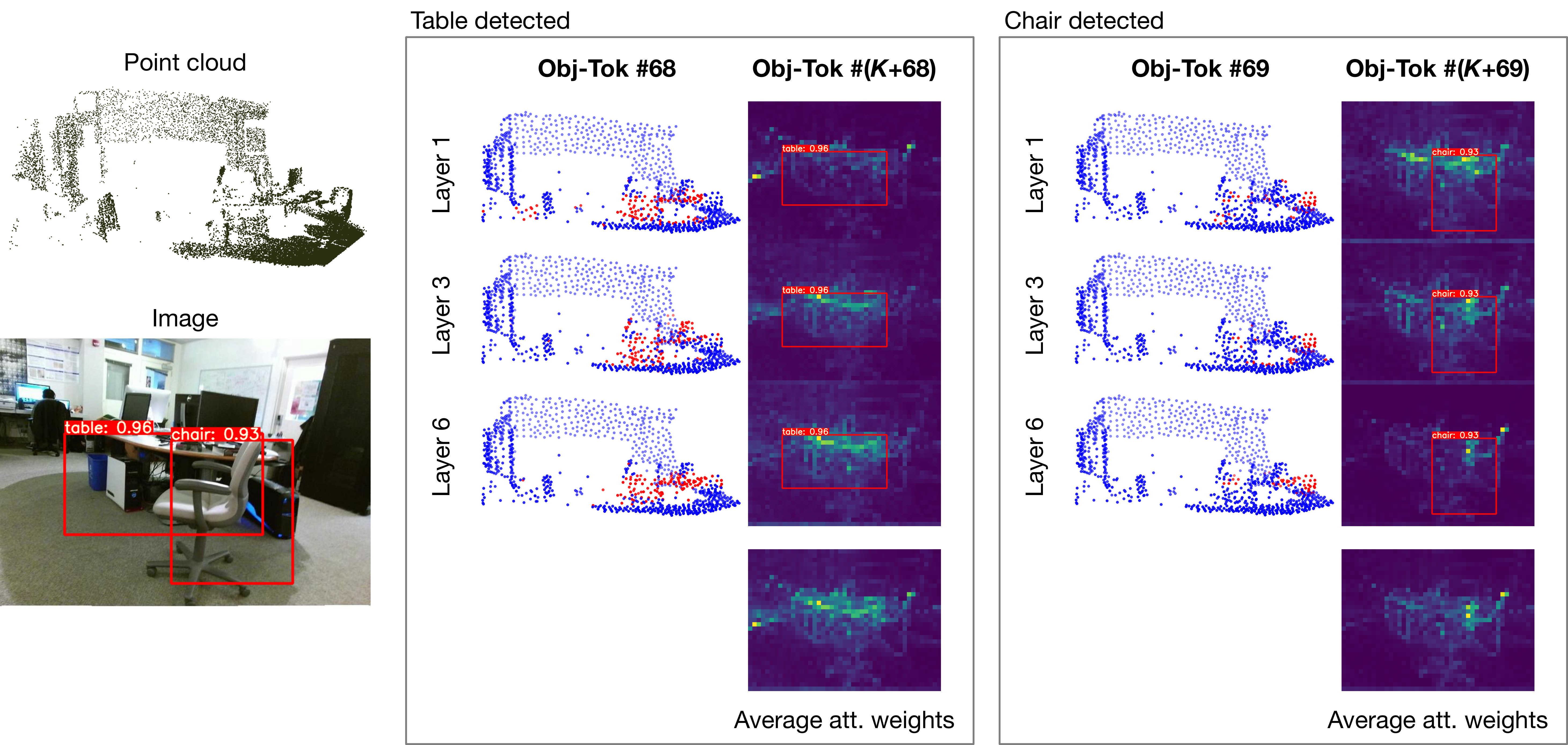}
\caption[]{Attention weights w.r.t. corresponding object queries of points and image patches. The points with larger  attention weights  than $2\times10^{-3}$  are colored as red, otherwise blue.}
\label{pic:weights}
\end{figure*}

The augmented extrinsic matrix after data augmentation is then calculated as $\bm{R}_t'=\bm{A}^\top \bm{R}_t$. 

As mentioned in our main paper, for ScanNetV2, we use depths to filter out the projected 3D points which should be occluded, but visible due to the sparsity of the point cloud.

\section{Additional Results and  Visualizations} 
In \cref{pic:additional_visual}, we visualize  the predicted 3D detection boxes on the ScanNetV2 to compare, and we also provide  predicted 2D  boxes on two example image views. 

\cref{pic:points_boxes} depicts the projections of 3D points to the corresponding 2D image views, where the 3D bounding boxes are treated as 8 points, which are projected to obtain 2D bounding boxes. 

In \cref{pic:weights}, we visualize the attention weights w.r.t. corresponding object queries of points and image patches. We observe that the $68$-th object token detects the table from 3D points, and the $(K+68)$-th object token detects table from 2D image patches. Similarly, the $69$-th and $(K+69)$-th object tokens both detect the chair from 3D points and 2D image patches, respectively. Such alignment demonstrates the effectiveness of our bridging technique by using conditional object queries (proposed in Sec.~\ref{sec:bridge_by_query}).

{\small
\bibliographystyle{ieee_fullname}
\bibliography{egbib}

\begin{thebibliography}{10}\itemsep=-1pt

\bibitem{journals/pami/BaltrusaitisAM19}
Tadas Baltrusaitis, Chaitanya Ahuja, and Louis{-}Philippe Morency.
\newblock Multimodal machine learning: {A} survey and taxonomy.
\newblock In {\em {IEEE} Trans. PAMI}, 2019.

\bibitem{carion2020end}
Nicolas Carion, Francisco Massa, Gabriel Synnaeve, Nicolas Usunier, Alexander
  Kirillov, and Sergey Zagoruyko.
\newblock End-to-end object detection with transformers.
\newblock In {\em ECCV}, 2020.

\bibitem{chen2020hierarchical}
Jintai Chen, Biwen Lei, Qingyu Song, Haochao Ying, Danny~Z Chen, and Jian Wu.
\newblock A hierarchical graph network for 3d object detection on point clouds.
\newblock In {\em CVPR}, 2020.

\bibitem{chen2017multi}
Xiaozhi Chen, Huimin Ma, Ji Wan, Bo Li, and Tian Xia.
\newblock Multi-view 3d object detection network for autonomous driving.
\newblock In {\em CVPR}, 2017.

\bibitem{DBLP:conf/cvpr/DaiCSHFN17}
Angela Dai, Angel~X. Chang, Manolis Savva, Maciej Halber, Thomas~A. Funkhouser,
  and Matthias Nie{\ss}ner.
\newblock Scannet: Richly-annotated 3d reconstructions of indoor scenes.
\newblock In {\em CVPR}, 2017.

\bibitem{dosovitskiy2020image}
Alexey Dosovitskiy, Lucas Beyer, Alexander Kolesnikov, Dirk Weissenborn,
  Xiaohua Zhai, Thomas Unterthiner, Mostafa Dehghani, Matthias Minderer, Georg
  Heigold, Sylvain Gelly, et~al.
\newblock An image is worth 16x16 words: Transformers for image recognition at
  scale.
\newblock {\em arXiv preprint arXiv:2010.11929}, 2020.

\bibitem{engelmann20203d}
Francis Engelmann, Martin Bokeloh, Alireza Fathi, Bastian Leibe, and Matthias
  Nie{\ss}ner.
\newblock 3d-mpa: Multi-proposal aggregation for 3d semantic instance
  segmentation.
\newblock In {\em CVPR}, 2020.

\bibitem{fang2021you}
Yuxin Fang, Bencheng Liao, Xinggang Wang, Jiemin Fang, Jiyang Qi, Rui Wu,
  Jianwei Niu, and Wenyu Liu.
\newblock You only look at one sequence: Rethinking transformer in vision
  through object detection.
\newblock {\em arXiv preprint arXiv:2106.00666}, 2021.

\bibitem{guo2020deep}
Yulan Guo, Hanyun Wang, Qingyong Hu, Hao Liu, Li Liu, and Mohammed Bennamoun.
\newblock Deep learning for 3d point clouds: A survey.
\newblock {\em T-PAMI}, 2020.

\bibitem{gwak2020generative}
JunYoung Gwak, Christopher Choy, and Silvio Savarese.
\newblock Generative sparse detection networks for 3d single-shot object
  detection.
\newblock In {\em arXiv preprint arXiv:2006.12356}, 2020.

\bibitem{kokkinos2012intrinsic}
Iasonas Kokkinos, Michael~M Bronstein, Roee Litman, and Alex~M Bronstein.
\newblock Intrinsic shape context descriptors for deformable shapes.
\newblock In {\em CVPR}, 2012.

\bibitem{lahoud20172d}
Jean Lahoud and Bernard Ghanem.
\newblock 2d-driven 3d object detection in rgb-d images.
\newblock In {\em CVPR}, 2017.

\bibitem{lin2014microsoft}
Tsung-Yi Lin, Michael Maire, Serge Belongie, James Hays, Pietro Perona, Deva
  Ramanan, Piotr Doll{\'a}r, and C~Lawrence Zitnick.
\newblock Microsoft coco: Common objects in context.
\newblock In {\em ECCV}, 2014.

\bibitem{liu2021swin}
Ze Liu, Yutong Lin, Yue Cao, Han Hu, Yixuan Wei, Zheng Zhang, Stephen Lin, and
  Baining Guo.
\newblock Swin transformer: Hierarchical vision transformer using shifted
  windows.
\newblock {\em arXiv preprint arXiv:2103.14030}, 2021.

\bibitem{liu2021group}
Ze Liu, Zheng Zhang, Yue Cao, Han Hu, and Xin Tong.
\newblock Group-free 3d object detection via transformers.
\newblock {\em arXiv preprint arXiv:2104.00678}, 2021.

\bibitem{DBLP:journals/corr/abs-1711-05101}
Ilya Loshchilov and Frank Hutter.
\newblock Fixing weight decay regularization in adam.
\newblock {\em arXiv preprint arXiv:1711.05101}, 2017.

\bibitem{mao2021voxel}
Jiageng Mao, Yujing Xue, Minzhe Niu, Haoyue Bai, Jiashi Feng, Xiaodan Liang,
  Hang Xu, and Chunjing Xu.
\newblock Voxel transformer for 3d object detection.
\newblock In {\em ICCV}, 2021.

\bibitem{meng2021conditional}
Depu Meng, Xiaokang Chen, Zejia Fan, Gang Zeng, Houqiang Li, Yuhui Yuan, Lei
  Sun, and Jingdong Wang.
\newblock Conditional detr for fast training convergence.
\newblock In {\em ICCV}, 2021.

\bibitem{misra2021end}
Ishan Misra, Rohit Girdhar, and Armand Joulin.
\newblock An end-to-end transformer model for 3d object detection.
\newblock In {\em ICCV}, 2021.

\bibitem{qi2020imvotenet}
Charles~R Qi, Xinlei Chen, Or Litany, and Leonidas~J Guibas.
\newblock Imvotenet: Boosting 3d object detection in point clouds with image
  votes.
\newblock In {\em CVPR}, 2020.

\bibitem{qi2019deep}
Charles~R Qi, Or Litany, Kaiming He, and Leonidas~J Guibas.
\newblock Deep hough voting for 3d object detection in point clouds.
\newblock In {\em ICCV}, 2019.

\bibitem{qi2018frustum}
Charles~R Qi, Wei Liu, Chenxia Wu, Hao Su, and Leonidas~J Guibas.
\newblock Frustum pointnets for 3d object detection from rgb-d data.
\newblock In {\em CVPR}, 2018.

\bibitem{qi2017pointnet}
Charles~R Qi, Hao Su, Kaichun Mo, and Leonidas~J Guibas.
\newblock Pointnet: Deep learning on point sets for 3d classification and
  segmentation.
\newblock In {\em CVPR}, 2017.

\bibitem{DBLP:conf/nips/QiYSG17}
Charles~R Qi, Li Yi, Hao Su, and Leonidas~J. Guibas.
\newblock Pointnet++: Deep hierarchical feature learning on point sets in a
  metric space.
\newblock In {\em NIPS}, 2017.

\bibitem{xie2020mlcvnet}
Xie Qian, Lai Yu-kun, Wu Jing, Wang Zhoutao, Zhang Yiming, Xu Kai, and Wang
  Jun.
\newblock Mlcvnet: Multi-level context votenet for 3d object detection.
\newblock In {\em CVPR}, 2020.

\bibitem{ren2015faster}
Shaoqing Ren, Kaiming He, Ross Girshick, and Jian Sun.
\newblock Faster r-cnn: Towards real-time object detection with region proposal
  networks.
\newblock {\em NIPS}, 2015.

\bibitem{song2015sun}
Shuran Song, Samuel~P Lichtenberg, and Jianxiong Xiao.
\newblock Sun rgb-d: A rgb-d scene understanding benchmark suite.
\newblock In {\em CVPR}, 2015.

\bibitem{song2016deep}
Shuran Song and Jianxiong Xiao.
\newblock Deep sliding shapes for amodal 3d object detection in rgb-d images.
\newblock In {\em CVPR}, 2016.

\bibitem{su2015multi}
Hang Su, Subhransu Maji, Evangelos Kalogerakis, and Erik Learned-Miller.
\newblock Multi-view convolutional neural networks for 3d shape recognition.
\newblock In {\em ICCV}, 2015.

\bibitem{pmlr-v139-touvron21a}
Hugo Touvron, Matthieu Cord, Matthijs Douze, Francisco Massa, Alexandre
  Sablayrolles, and Herve Jegou.
\newblock Training data-efficient image transformers and distillation through
  attention.
\newblock In {\em ICML}, 2021.

\bibitem{DBLP:conf/nips/WangHSXRH20}
Yikai Wang, Wenbing Huang, Fuchun Sun, Tingyang Xu, Yu Rong, and Junzhou Huang.
\newblock Deep multimodal fusion by channel exchanging.
\newblock In {\em NeurIPS}, 2020.

\bibitem{wu20153d}
Zhirong Wu, Shuran Song, Aditya Khosla, Fisher Yu, Linguang Zhang, Xiaoou Tang,
  and Jianxiong Xiao.
\newblock 3d shapenets: A deep representation for volumetric shapes.
\newblock In {\em CVPR}, 2015.

\bibitem{xie2021segformer}
Enze Xie, Wenhai Wang, Zhiding Yu, Anima Anandkumar, Jose~M Alvarez, and Ping
  Luo.
\newblock Segformer: Simple and efficient design for semantic segmentation with
  transformers.
\newblock In {\em NeurIPS}, 2021.

\bibitem{xu2018pointfusion}
Danfei Xu, Dragomir Anguelov, and Ashesh Jain.
\newblock Pointfusion: Deep sensor fusion for 3d bounding box estimation.
\newblock In {\em CVPR}, 2018.

\bibitem{yuan2021hrformer}
Yuhui Yuan, Rao Fu, Lang Huang, Weihong Lin, Chao Zhang, Xilin Chen, and
  Jingdong Wang.
\newblock Hrformer: High-resolution transformer for dense prediction.
\newblock {\em arXiv preprint arXiv:2110.09408}, 2021.

\bibitem{zhang2020h3dnet}
Zaiwei Zhang, Bo Sun, Haitao Yang, and Qixing Huang.
\newblock H3dnet: 3d object detection using hybrid geometric primitives.
\newblock {\em arXiv preprint arXiv:2006.05682}, 2020.

\bibitem{SETR}
Sixiao Zheng, Jiachen Lu, Hengshuang Zhao, Xiatian Zhu, Zekun Luo, Yabiao Wang,
  Yanwei Fu, Jianfeng Feng, Tao Xiang, Philip~H.S. Torr, and Li Zhang.
\newblock Rethinking semantic segmentation from a sequence-to-sequence
  perspective with transformers.
\newblock In {\em CVPR}, 2021.

\bibitem{zhou2018voxelnet}
Yin Zhou and Oncel Tuzel.
\newblock Voxelnet: End-to-end learning for point cloud based 3d object
  detection.
\newblock In {\em CVPR}, 2018.

\bibitem{zhu2020deformable}
Xizhou Zhu, Weijie Su, Lewei Lu, Bin Li, Xiaogang Wang, and Jifeng Dai.
\newblock Deformable detr: Deformable transformers for end-to-end object
  detection.
\newblock {\em arXiv preprint arXiv:2010.04159}, 2020.

\end{thebibliography}
}

\end{document}